\documentclass{article}

\usepackage{microtype}
\usepackage{graphicx}
\usepackage{subcaption}
\usepackage{booktabs}
\usepackage{hyperref}
\usepackage{xurl}      
\usepackage{pgfplots}
\usepackage{tikz}
\usepackage{xcolor}

\usepackage[preprint]{icml2026}

\makeatletter
\renewcommand{\Notice@String}{Preliminary work. Under review at the 5th DL4C Workshop @ ICML 2026.}
\makeatother

\usepackage{amsmath}
\usepackage{amssymb}

\pgfplotsset{compat=1.18}

\definecolor{goosecolor}{RGB}{52, 152, 219}
\definecolor{opencodecolor}{RGB}{231, 76, 60}
\definecolor{openhandscolor}{RGB}{46, 204, 113}
\definecolor{qwencolor}{RGB}{155, 89, 182}
\definecolor{minimaxcolor}{RGB}{230, 126, 34}

\icmltitlerunning{The Scaffold Effect in Coding Agents}

\begin{document}

\twocolumn[
  \icmltitle{The Scaffold Effect in Coding Agents:\\
    Harness Choice as a Hidden Variable in Coding-Agent Evaluation}

  \begin{icmlauthorlist}
    \icmlauthor{Naman Vats}{sl}
    \icmlauthor{Oleg Golev}{sl}
  \end{icmlauthorlist}

  \icmlaffiliation{sl}{Sentient Labs}

  \icmlcorrespondingauthor{Naman Vats, Oleg Golev}{\{naman, oleg\}@sentient.xyz}

  \icmlkeywords{coding agents, evaluation, harness, scaffold, terminal-bench, human-centered AI}

  \vskip 0.3in
]

\printAffiliationsAndNotice{}

\begin{abstract}
Public leaderboards for coding agents typically rank systems by model name and pass rate, while the surrounding \emph{harness}---the scaffold that issues tools, manages context, and decides when to stop---is often under-specified. Model-to-model comparison is valid when the harness is fixed; when it varies, performance and efficiency conflate model and scaffold effects. We evaluate Qwen~3.6~Plus and MiniMax~M2.5 across three open-source harnesses (Goose, OpenCode, OpenHands-SDK) on a stratified 50-task subset of Terminal-Bench Pro. Harness choice induces up to a $\mathbf{40\times}$ difference in tokens per solved task, while paired within-model pass-rate differences remain 0--8 percentage points (95\% paired-task bootstrap CIs include zero except for the largest gap). Failure fingerprints replicate across models (REASON for Goose, VERIFY/MAX\_TURNS for OpenHands-SDK, idle-loop/TIME for OpenCode), indicating harness-level biases that are largely model-independent. For \emph{human-centered coding-agent evaluation}, model name alone is an incomplete comparison unit: harness--model pairs determine real-world cost, latency, and oversight burden---no-action turns are a per-task wait tax, not just a token tax. We therefore recommend selecting harness--model pairs by pass rate under token/latency budgets, and reporting token usage, latency, and full harness specifications alongside any model comparison. We release anonymized configs, raw trial logs, aggregated snapshots, and analysis scripts.
\end{abstract}

\section{Introduction}
\label{sec:intro}

Coding-agent leaderboards have become the dominant benchmark for tracking progress in AI software development. They rank entries by model name, with the surrounding agent harness either undisclosed, varied to maximize per-row score, or implicitly treated as a controlled constant.
From a \emph{human-centered} view of coding agents, this raises a pragmatic concern: a developer choosing an agent for daily work cares about cost per resolved task, time-to-completion, and how much oversight is needed to keep the agent productive. None of those are recovered from a model-only score.

The intuition that harness matters is not new. The Terminal-Bench~2.0 paper~\citep{merrill2026terminalbench} reports Claude~Opus~4.5 at 52.1\% with one harness and 57.8\% with another, while consuming 256.9M versus 3.9M input tokens respectively, a 65$\times$ token difference for a 5.7-point accuracy gain.
This paper systematizes that observation into a controlled study and asks: when the model is held fixed and only the harness varies, how large are the differences a deployer should expect, and what kinds of differences are they?

\paragraph{Contributions.}
\begin{enumerate}
  \item Across 300 trials (3 harnesses $\times$ 2 agentic coding models $\times$ 50 Terminal-Bench Pro tasks), we measure a $\mathbf{40\times}$ gap in tokens per solved task; paired pass-rate differences stay within 0--8 pp (95\% paired-task bootstrap CIs include zero for all but the largest pairwise gap).
  \item We show the harness-specific failure-mode fingerprint replicates across both models, identifying it as a scaffold property rather than a model property.
  \item We provide a mechanistic proxy linking \emph{no-action turns} (turns in which the agent neither edits a file nor issues a new command) to the cost gap, and frame these as a per-task \emph{oversight tax} that a human-in-the-loop user must absorb.
  \item We argue the unit of comparison for human-centered coding-agent evaluation should be the \emph{harness--model pair}, not the model alone, and include anonymized configs and raw trial data as supplementary material.\footnote{\url{https://anonymous.4open.science/r/scaffold-effects-dl4c-supp/}}
\end{enumerate}

\section{Related Work}
\label{sec:related}

\paragraph{Coding-agent benchmarks.}
Terminal-Bench~\citep{merrill2026terminalbench} evaluates agents on realistic command-line tasks requiring compilation, environment setup, and iterative debugging.
Terminal-Bench Pro~\citep{wang2025tbpro,tbpro_dataset}\footnote{We follow the dataset-requested citation listed on the Terminal-Bench Pro Hugging Face card~\citep{wang2025tbpro}; the benchmark is introduced as a component of that broader paper. The benchmark artifact itself is released by Alibaba at \url{https://github.com/alibaba/terminal-bench-pro}~\citep{tbpro_dataset}.} extends this to 400 tasks across 8 domains with higher test coverage ($\sim$28 tests per task).
SWE-bench~\citep{jimenez2024swebench} and successors note that scaffold choice affects scores but treat it as a confound to control rather than a phenomenon to study.
Our 50-task subset of Terminal-Bench Pro covers 8 archetypes (BUG, BUILD, DATA, IMPL, ML, PUZZLE, SEC, SYS).

\paragraph{Agent scaffolding and evaluation reliability.}
\citet{wang2024agentsystem} argue that agents are systems rather than models, but do not quantify the efficiency impact across harnesses on a fixed task set with a fixed model.
The OpenHands platform paper~\citep{openhandssdk2025} introduces a composable SDK for production software-engineering agents. Goose~\citep{goose2026} (originally Block, Inc., now Agentic AI Foundation) and OpenCode~\citep{opencode2025} are widely deployed open-source coding agents.
Harbor~\citep{harbor2026} provides a unified evaluation framework supporting these harnesses with a single task manifest, which we use as our execution infrastructure.

\paragraph{Cost, oversight, and human-centered evaluation.}
Token efficiency has been studied for long-context benchmarks~\citep{hsieh2024ruler} but not systematically for agentic coding evaluation.
The DL4C call for ``Interaction-Aware Benchmarks'' explicitly asks for metrics beyond task completion that capture interaction quality, oversight burden, and verifiability.
Our study contributes a controlled per-task measurement of these properties: tokens, turns, idle time, and failure-mode mix.

\section{Experimental Setup}
\label{sec:setup}

\subsection{Tasks}
We use Terminal-Bench Pro~\citep{wang2025tbpro,tbpro_dataset}, selecting 50 tasks from the 200-task public set via stratified random sampling across 8 domain categories (Table~\ref{tab:tasks}). All tasks have deterministic evaluation via \texttt{pytest} suites. Tasks were first categorized by GPT-4o-mini classification of each task's \texttt{instruction.md}; per-category counts follow the source-set distribution, and tasks within each category were drawn randomly.

\begin{table}[t]
\centering
\caption{Task distribution across 8 archetypes ($n=50$).}
\label{tab:tasks}
\small
\begin{tabular}{lcl}
\toprule
\textbf{Cat.} & \textbf{Count} & \textbf{Description} \\
\midrule
BUG    & 8 & Bug fix / debug \\
BUILD  & 6 & Build / compile / env setup \\
DATA   & 7 & Data transformation / parsing \\
IMPL   & 8 & Fresh implementation \\
ML     & 8 & ML / model training / eval \\
PUZZLE & 5 & Puzzle / game solving \\
SEC    & 6 & Security / forensics / recovery \\
SYS    & 2 & System / service configuration \\
\bottomrule
\end{tabular}
\end{table}

\subsection{Models}
Two recent strong agentic coding models, accessed through the OpenRouter API:
\begin{itemize}
  \item \textbf{Qwen 3.6 Plus} (Alibaba Cloud): hybrid linear-attention MoE with always-on chain-of-thought; reported 61.6 on Terminal-Bench~2.0 versus Claude~Opus~4.5 at 59.3~\citep{qwen36plus2026}.
  \item \textbf{MiniMax M2.5}: 228.7B-parameter MoE, 10B active; reported 80.2\% on SWE-bench Verified~\citep{minimax2026}.
\end{itemize}

\subsection{Harnesses}
Three open-source coding-agent harnesses with distinct scaffolding philosophies:
\begin{itemize}
  \item \textbf{Goose}~\citep{goose2026}: heavyweight IDE agent with eager file-tree pre-injection.
  \item \textbf{OpenCode}~\citep{opencode2025}: persistent tool-loop coding agent; no automatic context pre-loading.
  \item \textbf{OpenHands-SDK}~\citep{openhandssdk2025}: micro-agent architecture with sub-agent delegation, internal retry, and explicit verification steps.
\end{itemize}
All three are natively supported by Harbor and were run with identical task manifests.

\subsection{Standardization Protocol}
\label{sec:protocol}
Held constant: verbatim Terminal-Bench Pro instruction; native test suite; Daytona sandbox per task; 900-second wall-time cap per trial; OpenRouter as the model gateway. The \emph{system prompt template} was held \emph{largely identical} across the three harnesses: same operating principles, same non-negotiables, the same shared four-skill operational playbook (\texttt{build-and-env}, \texttt{deep-debug}, \texttt{terminal-investigation}, \texttt{test-driven-solve}), and the same task instruction. Only a short harness-specific ``Using your tools'' paragraph and turn/iteration vocabulary differ.
\textbf{Maximum turns:} 40 for Goose and OpenHands-SDK (passed through agent kwargs); OpenCode does not expose a turn-budget flag through Harbor and is bounded only by the wall-time cap. We intentionally evaluate each harness as a deployable system under its native control surface; \emph{exposed} budget controls are part of the scaffold being studied, not nuisance parameters to equalize.
What was \emph{not} standardized, and constitutes the phenomenon under study, is each harness's native tool API, automatic context pre-loading, and internal retry/sub-agent logic.

\subsection{Metrics}
\label{sec:metrics}
For each trial we collect: \texttt{solved} (bool), \texttt{turns\_used}, \texttt{tokens\_total}, \texttt{avg\_no\_action\_turns} (turns where no file was modified \emph{and} no new shell command was issued; we treat this as a \emph{proxy} for oversight burden, since a reasoning-only turn could in principle still be useful), \texttt{hit\_turn\_budget\_count}, \texttt{wall\_seconds}, and a post-hoc \texttt{failure\_category} from a 6-class taxonomy (REASON, VERIFY, TIME, MAX\_TURNS, HANG, ERROR; classification rule in Appendix~\ref{app:failure}). Bootstrap uncertainty is reported as 95\% paired-task bootstrap intervals ($B=10{,}000$) for both pass rate and tokens per solved task.

\paragraph{Token accounting.}
Tokens come from a layered fallback in our analysis pipeline. For OpenCode and OpenHands-SDK we use the harness-emitted ATIF total of prompt + completion + cached tokens (input/output split available). For Goose we use the harness's total-only field (Goose does not expose a per-direction split through Harbor's standard interface). \texttt{tokens\_total} is the sum the caller is billed for; we do not normalize across providers because the harness-reported number is the deployment-relevant quantity. \emph{Tokens per solved task is amortized}: the cell-wide sum of \texttt{tokens\_total} divided by the cell's solved count, $\sum_{t\in\text{cell}} \texttt{tokens\_total}_t / |\{t : \texttt{solved}_t\}|$. Failed and infrastructure-error trials \emph{are} included in the numerator (their tokens count); HANG and ERROR trials that exit before any LLM call simply contribute 0. This is the cost a deployer absorbs per successful task, not the cost of a successful trial alone.

\paragraph{Provider settings.} All trials route through the OpenRouter API. We use each model's default OpenRouter sampling (no explicit \texttt{temperature}, \texttt{top\_p}, or \texttt{max\_tokens} override; whichever defaults the provider applies); no provider-routing override; no harness-level retries on tool-call failure. Infrastructure failures (sandbox quota, image build) are classified as ERROR and counted in the denominator of the cell ($n=50$) but not as solved.

\paragraph{Task selection.} The 50 tasks were drawn from the 200-task public set \emph{before} any harness or model was run; no task was removed after results were observed. Categorization used GPT-4o-mini classification of each task's \texttt{instruction.md}. The full categorization output and the 50 task IDs are included as supplementary material.

\section{Results}
\label{sec:results}

\subsection{Pass Rate}
\label{sec:passrate}
Table~\ref{tab:passrate} reports per-cell pass rates with 95\% paired-task bootstrap CIs. Within a model, the harness range is 2--8~pp; across models within a harness it is 4--10~pp. The pairwise differences with their CIs are: at most $-8.0$~pp (Goose vs.\ OpenCode on MiniMax, 95\% CI $[-18.0, 0.0]$), $-2.0$~pp (Goose vs.\ OpenCode on Qwen, $[-12.0, +8.0]$), and $0.0$~pp between OpenCode and OpenHands-SDK on either model.
\textbf{At $n=50$, most pairwise pass-rate differences are not statistically distinguishable from zero.}

\begin{table}[t]
\centering
\caption{Pass rates with 95\% paired-task bootstrap CIs ($B{=}10{,}000$, $n=50$ per cell).}
\label{tab:passrate}
\small
\begin{tabular}{lcc}
\toprule
\textbf{Harness} & \textbf{Qwen 3.6 Plus} & \textbf{MiniMax M2.5} \\
\midrule
Goose         & 48.0\% [34.0, 62.0]  & 38.0\% [24.0, 52.0] \\
OpenCode      & 50.0\% [36.0, 64.0]  & 46.0\% [32.0, 60.0] \\
OpenHands-SDK & 50.0\% [36.0, 64.0]  & 46.0\% [32.0, 60.0] \\
\bottomrule
\end{tabular}
\end{table}

\subsection{Token Cost per Solved Task}
Table~\ref{tab:tokens} reports tokens per solved task, the primary efficiency metric. The ordering is identical for both models: Goose $\ll$ OpenHands-SDK $<$ OpenCode. OpenCode consumes approximately 40$\times$ more tokens per solved task than Goose, while pass-rate differences remain within the 0--8~pp harness range above. The gap is \emph{not} driven by OpenCode running more turns: average turn counts are 21--27 for OpenCode versus 18--25 for Goose ($\sim$1.2$\times$, well within Goose's 40-turn cap). The per-task token gap is therefore not primarily attributable to turn-count differences; it more likely reflects per-turn context growth, tool serialization, and harness-specific token accounting.

\paragraph{Bootstrap robustness.} 95\% paired-task bootstrap CIs ($B{=}10{,}000$) on tokens per solved task are: Goose~Qwen [21K, 40K]; Goose~MM [25K, 61K]; OpenHands~Qwen [610K, 1.21M]; OpenHands~MM [537K, 1.35M]; OpenCode~Qwen [733K, 1.84M]; OpenCode~MM [1.01M, 2.50M]. Goose's CI upper bound (40--61K) sits well below OpenCode's CI lower bound (733K--1.01M), so the order-of-magnitude gap is robust to bootstrap uncertainty.

\paragraph{OpenCode budget-control sensitivity.} Because OpenCode lacks a turn-budget flag, we check whether the cost gap could be an artefact of long runs. Of OpenCode trials with recorded turn counts, 88\% (Qwen, 42/48) and 71\% (MiniMax, 34/48) stayed below 40 turns; the maximum observed turn count was 67--69. A 40-turn cap (matching Goose and OpenHands-SDK) would therefore truncate $\sim$12--29\% of OpenCode trials, mostly already in the failure tail. The 40$\times$ token gap does not come from OpenCode running unboundedly long; it comes from per-turn token volume.

\begin{table}[t]
\centering
\caption{Tokens per solved task and average turns. Lower is better.}
\label{tab:tokens}
\small
\setlength{\tabcolsep}{4pt}
\begin{tabular}{llrrr}
\toprule
\textbf{Harness} & \textbf{Model} & \textbf{Tok/Solve} & \textbf{$\times$Goose} & \textbf{Avg.\ Turns} \\
\midrule
Goose     & Qwen     & 28{,}142    & $1.0$  & 17.96 \\
Goose     & MiniMax  & 36{,}950    & $1.0$  & 25.25 \\
\addlinespace[1pt]
OpenHands & Qwen     & 841{,}201   & $29.9$ & 25.95 \\
OpenHands & MiniMax  & 843{,}286   & $22.8$ & 24.19 \\
\addlinespace[1pt]
OpenCode  & Qwen     & 1{,}147{,}740 & $40.8$ & 21.71 \\
OpenCode  & MiniMax  & 1{,}546{,}977 & $41.9$ & 27.46 \\
\bottomrule
\end{tabular}
\end{table}

\subsection{Pareto Frontier: Pass Rate vs.\ Token Cost}
\label{sec:pareto}

\begin{figure}[t]
\centering
\begin{tikzpicture}
\begin{axis}[
    width=0.92\columnwidth,
    height=5.8cm,
    xlabel={Tokens per solved task (log scale)},
    ylabel={Pass rate (\%)},
    xmode=log,
    xmin=18000, xmax=3000000,
    ymin=30, ymax=58,
    grid=both, grid style={dashed, gray!30},
    legend style={at={(0.98,0.04)}, anchor=south east, font=\scriptsize, draw=gray!40, fill=white, fill opacity=0.9, draw opacity=1, text opacity=1},
    legend cell align={left},
    label style={font=\small},
    tick label style={font=\scriptsize},
]
\addplot[dashed, gray, thick, no marks, forget plot] coordinates {
    (28142, 48) (841201, 50)
};
\addplot[only marks, mark=*, mark size=3pt, color=goosecolor] coordinates {
    (28142, 48) (36950, 38)
};
\addlegendentry{Goose}
\addplot[only marks, mark=square*, mark size=2.6pt, color=openhandscolor] coordinates {
    (841201, 50) (843286, 46)
};
\addlegendentry{OpenHands-SDK}
\addplot[only marks, mark=triangle*, mark size=3.4pt, color=opencodecolor] coordinates {
    (1147740, 50) (1546977, 46)
};
\addlegendentry{OpenCode}
\end{axis}
\end{tikzpicture}
\caption{Pareto plot: pass rate (y) vs.\ tokens per solved task (x, log). Goose dominates the frontier; OpenCode is Pareto-dominated for both models. The pass-rate spread is small (38--50\%); the cost spread is two orders of magnitude.}
\label{fig:pareto}
\end{figure}
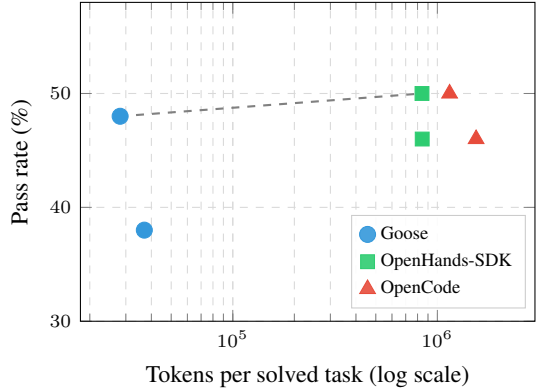

Figure~\ref{fig:pareto} plots both axes on the same panel. Goose lies on the Pareto frontier for both models. No point dominates it on cost-at-equal-or-better-pass-rate. OpenCode is Pareto-dominated by both Goose (on cost, with comparable pass rate) and OpenHands-SDK (on cost, with the same pass rate).

\subsection{No-Action Turns}
\label{sec:noprogress}

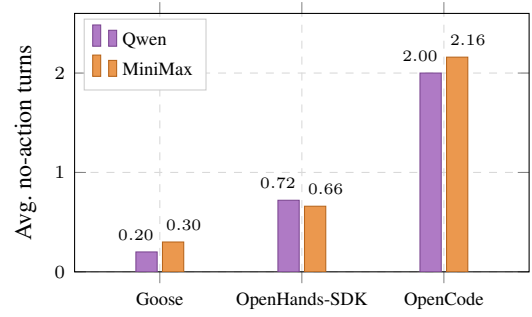
\begin{figure}[t]
\centering
\begin{tikzpicture}
\begin{axis}[
    ybar,
    bar width=8pt,
    width=0.92\columnwidth,
    height=5cm,
    ylabel={Avg.\ no-action turns},
    symbolic x coords={Goose, OpenHands-SDK, OpenCode},
    xtick=data,
    xticklabel style={font=\scriptsize},
    ymin=0, ymax=2.6,
    legend style={at={(0.02,0.98)}, anchor=north west, font=\scriptsize, draw=gray!40},
    legend cell align={left},
    grid=both, grid style={dashed, gray!30},
    label style={font=\small},
    tick label style={font=\scriptsize},
    nodes near coords,
    every node near coord/.append style={font=\tiny, /pgf/number format/fixed, /pgf/number format/precision=2, /pgf/number format/fixed zerofill, anchor=south, yshift=1pt},
    enlarge x limits=0.3,
]
\addplot[fill=qwencolor!80, draw=qwencolor!80!black,
    every node near coord/.append style={xshift=-4pt}
] coordinates {
    (Goose, 0.20) (OpenHands-SDK, 0.72) (OpenCode, 2.00)
};
\addplot[fill=minimaxcolor!80, draw=minimaxcolor!80!black,
    every node near coord/.append style={xshift=4pt}
] coordinates {
    (Goose, 0.30) (OpenHands-SDK, 0.66) (OpenCode, 2.16)
};
\legend{Qwen, MiniMax}
\end{axis}
\end{tikzpicture}
\caption{Average no-action turns per task. OpenCode exhibits 10$\times$ more no-action turns than Goose, regardless of model. This pattern is consistent with the 40$\times$ token gap through compounding context accumulation.}
\label{fig:nop}
\end{figure}

OpenCode averages 2.0--2.16 no-action turns per task versus 0.2--0.3 for Goose, a 10$\times$ ratio that replicates across both models. Each no-action turn carries a full context window of input tokens due to message-history accumulation, so two such turns per task at average positions $\sim 10$ and $\sim 20$ each waste thousands of tokens. We frame this as the per-task \emph{idle-turn tax}: a developer running OpenCode interactively absorbs both the dollar cost and the wall-clock latency of these idle loops.

\subsection{Failure-Mode Fingerprints}
\label{sec:failures}

Table~\ref{tab:failures} reports the 6-class failure breakdown per cell. Three distinct fingerprints emerge that replicate across both models:

\begin{itemize}
  \item \textbf{Goose: REASON-dominated.} 20/15 REASON failures (Qwen/MiniMax) with zero VERIFY. When stuck, Goose stops rather than commit to a wrong solution. The low \texttt{avg\_no\_action\_turns} (0.2/0.3) confirms decisive action.
  \item \textbf{OpenHands-SDK: VERIFY + MAX\_TURNS.} 6/8 VERIFY and 6/6 MAX\_TURNS. \texttt{hit\_turn\_budget\_count} is 8/9, exceeding the MAX\_TURNS classification of 6/6 because runs that hit the budget while also raising an exception are routed to HANG by the decision tree (Appendix~\ref{app:failure}).
  \item \textbf{OpenCode: TIME + idle spinning.} 5/10 TIME and 4/1 HANG, with zero VERIFY. OpenCode shows 0/0 MAX\_TURNS because the harness lacks a turn-budget flag through Harbor (Section~\ref{sec:protocol}); runs that would manifest as MAX\_TURNS under Goose or OpenHands-SDK instead surface here as TIME or HANG.
\end{itemize}

\begin{table}[t]
\centering
\caption{Failure-category counts per cell, $n=50$. Columns:
\textbf{Sv}=solved, \textbf{Rs}=REASON, \textbf{Vf}=VERIFY,
\textbf{Tm}=TIME, \textbf{MT}=MAX\_TURNS, \textbf{Hg}=HANG, \textbf{Er}=ERROR.}
\label{tab:failures}
\small
\setlength{\tabcolsep}{3.5pt}
\begin{tabular}{llrrrrrrr}
\toprule
\textbf{Harness} & \textbf{Model} & \textbf{Sv} & \textbf{Rs} & \textbf{Vf} & \textbf{Tm} & \textbf{MT} & \textbf{Hg} & \textbf{Er} \\
\midrule
Goose     & Qwen    & 23 & 20 & 0 &  3 & 1 & 2 & 1 \\
Goose     & MiniMax & 17 & 15 & 0 &  7 & 8 & 2 & 1 \\
OpenCode  & Qwen    & 25 & 15 & 0 &  5 & 0 & 4 & 1 \\
OpenCode  & MiniMax & 22 & 16 & 0 & 10 & 0 & 1 & 1 \\
OpenHands & Qwen    & 25 &  3 & 6 &  5 & 6 & 4 & 1 \\
OpenHands & MiniMax & 21 &  1 & 8 &  6 & 6 & 7 & 1 \\
\bottomrule
\end{tabular}
\end{table}

\subsection{Pass Rate by Task Category}
\label{sec:bycategory}
Table~\ref{tab:bycat} pools both models per cell and reports pass rate per (category, harness). Differences are mostly small, but one is conspicuous: on \textbf{IMPL} (fresh implementation), OpenHands-SDK substantially outperforms Goose and OpenCode (69\% vs.\ 38\%/38\%). All three harnesses struggle on \textbf{SEC} (8--17\%) and fail completely on \textbf{SYS} (0\%). For the remaining categories, harness rank order is unstable across categories, a useful caveat for any per-category leaderboard.

\begin{table}[t]
\centering
\caption{Pass rate by task category (pooled over both models, $n=2{\times}|\text{cat}|$ trials per harness).}
\label{tab:bycat}
\small
\setlength{\tabcolsep}{4pt}
\begin{tabular}{lrrrr}
\toprule
\textbf{Cat.} & \textbf{\#Tasks} & \textbf{Goose} & \textbf{OpenCode} & \textbf{OpenHands} \\
\midrule
BUG    & 8 & 62\% & 75\% & 62\% \\
BUILD  & 6 & 50\% & 50\% & 42\% \\
DATA   & 7 & 57\% & 71\% & 64\% \\
IMPL   & 8 & 38\% & 38\% & \textbf{69}\% \\
ML     & 8 & 56\% & 56\% & 44\% \\
PUZZLE & 5 & 30\% & 30\% & 40\% \\
SEC    & 6 &  8\% & 17\% & 17\% \\
SYS    & 2 &  0\% &  0\% &  0\% \\
\bottomrule
\end{tabular}
\end{table}

\subsection{Summary of Harness vs.\ Model Effects}
\begin{table}[t]
\centering
\caption{Comparative impact of harness choice vs.\ model upgrade.}
\label{tab:summary}
\footnotesize
\setlength{\tabcolsep}{3pt}
\begin{tabular}{lcc}
\toprule
\textbf{Variable} & \textbf{Harness} & \textbf{Model upgrade} \\
\midrule
Pass rate (paired)     & 0--8 pp     & 4--10 pp \\
Tokens/solved task     & \textbf{40$\times$}   & 1.0--1.3$\times$ \\
Failure profile        & \textbf{fingerprint} & consistent \\
No-action turns      & \textbf{10$\times$}   & $<$1.1$\times$ \\
\bottomrule
\end{tabular}
\end{table}

Pass-rate effects of harness and model upgrade are of comparable magnitude and within bootstrap noise at $n=50$. The cost-side asymmetry is overwhelming: harness choice shifts tokens-per-solved-task by 40$\times$, while upgrading the model barely moves it (1.0--1.3$\times$).

\section{Discussion}
\label{sec:discussion}

\subsection{What the Cost Gap Means for a Human User}
The 40$\times$ token gap is not just an accounting artefact; it has direct consequences for the developer who runs these agents in the loop:
\begin{itemize}
  \item \textbf{Dollar cost.} For one model and one task, $\sim$40$\times$ more tokens means $\sim$40$\times$ more API spend, all else equal. This dominates the 1--2$\times$ price spread between the strongest current models.
  \item \textbf{Wall-clock time.} OpenCode's no-action turns are not silent: each one is a round-trip API call the user waits through. With $\sim$2 idle turns per task, the user pays a per-task \emph{wait tax} on top of the dollar tax.
  \item \textbf{Oversight burden.} A user supervising the agent has to read or skim the idle turns to confirm they were not destructive. ``Re-reading the same file three times'' is cheap for the model but expensive for the human reviewer.
\end{itemize}

\subsection{What the Failure Fingerprints Mean}
The three harnesses have systematically different failure modes regardless of which model they run:
\begin{itemize}
  \item Goose stops cleanly when stuck (REASON-dominated, zero VERIFY). A user gets an honest ``I cannot do this'' rather than a plausible-but-wrong patch.
  \item OpenHands-SDK persists toward closure, accepting plausible-but-incorrect solutions (VERIFY) or running out of iterations (MAX\_TURNS). A user must then verify outputs more carefully.
  \item OpenCode commits only to solutions that pass tests (zero VERIFY), but more often exhausts the wall-time cap (TIME, HANG). A user pays in time and tokens, but the agent's positive verdicts can be trusted.
\end{itemize}
For human-centered coding-agent evaluation, this is a richer signal than pass rate alone: each fingerprint implies a different oversight discipline.

\subsection{Implications for Benchmark Reporting}
We recommend that coding-agent leaderboards adopt \textbf{harness--model pairs} as the unit of evaluation, with three first-class metrics alongside pass rate: tokens per solved task, average no-action turns per task, and the failure-category vector. Reporting only pass rate against model name conflates two independent sources of variance and discards the cost and oversight signals that determine whether a coding agent is actually deployable.

\section{Limitations}
\label{sec:limitations}

\begin{itemize}
\item \textbf{Sample size.} $n=50$ tasks; with one task corresponding to 2~percentage points, several pass-rate effects we observe are within bootstrap noise. We treat them as descriptive rather than statistically definitive.

\item \textbf{Token accounting asymmetry.} Goose totals are reported without an input/output breakdown via Harbor's standard interface. The 40$\times$ ratio is robust to this: it is anchored in totals, which Goose does report. Per-direction comparisons, however, are not available for Goose.

\item \textbf{Turn-budget asymmetry.} Goose and OpenHands-SDK enforce a 40-turn cap; OpenCode is bounded only by the 900-second wall-time cap, because OpenCode does not expose a turn-budget flag through Harbor. ``Turn'' is also harness-defined and may include differing numbers of internal LLM calls per harness.

\item \textbf{Three harnesses, two models.} The harness-specific findings should not be assumed to generalize to all scaffold architectures or to commercial closed-source agent products that we did not evaluate.

\item \textbf{Provider defaults may evolve.} OpenRouter and the upstream provider defaults (sampling parameters, route selection, model version) can shift between runs; exact replication requires pinning the provider route and sampling settings that were in effect at our run time. The released run metadata records the settings available at execution time.
\end{itemize}

\section{Conclusion}
\label{sec:conclusion}

We present controlled evidence that harness choice introduces a $\mathbf{40\times}$ token-cost difference and harness-specific failure fingerprints in coding-agent evaluation, while shifting paired pass rate by at most 8~percentage points. Both replicate independently across two recent models. For the human-centered coding-agent agenda, this argues that the unit of evaluation should be the \emph{harness--model pair}, with cost, idle-turn, and failure-mode signals reported alongside pass rate. We include anonymized configs, raw trial logs, aggregated snapshots, and analysis scripts as supplementary material.

\section*{Data and Code Availability}
Anonymized harness configurations, raw Harbor trial logs, aggregated snapshots, and the analysis pipeline are available at \url{https://anonymous.4open.science/r/scaffold-effects-dl4c-supp/}, frozen for the duration of review.

\section*{Impact Statement}
This paper presents work whose goal is to advance the field of Machine Learning, specifically the evaluation of coding agents. By showing that scaffold choice drives a $40\times$ token-cost gap and harness-specific failure profiles, we hope to encourage benchmark reporting that surfaces deployment cost and oversight burden, which directly affects the dollar cost, latency, and supervisory effort a human developer absorbs when using these systems. We see no specific ethical concern beyond those generally associated with releasing evaluation tooling and trial logs for coding agents.

\bibliographystyle{icml2026}
\bibliography{scaffold-effects}

\appendix

\section{Failure Classification Decision Tree}
\label{app:failure}
Failure classification is exception-first and matches the analysis pipeline (\texttt{analysis/collect.py},\penalty0\ \texttt{\_classify\_failure}). The ordered rules are:
\begin{enumerate}
  \item If \texttt{exception\_type} is non-null, map by exception class:
    \texttt{AgentTimeoutError}\,$\to$\,\textbf{TIME} (the 900-second per-trial \texttt{agent.timeout\_sec} fired);
    \texttt{NonZeroAgentExitCodeError},
    \texttt{AgentSetupError}\,$\to$\,\textbf{HANG};
    \texttt{DaytonaAuthorizationError},
    \texttt{EnvironmentBuildError},
    \texttt{RewardFileNotFoundError},
    parse errors\,$\to$\,\textbf{ERROR}.
  \item Else if \texttt{solved}$=$true\,$\to$\,\textbf{SOLVED}.
  \item Else if \texttt{hit\_turn\_budget}$=$true\,$\to$\,\textbf{MAX\_TURNS}. (\texttt{hit\_turn\_budget} is \texttt{turns\_used}$\geq$\texttt{max\_turns}; only Goose and OpenHands-SDK pass a turn cap to Harbor, so OpenCode never produces this category.)
  \item Else if \texttt{finish\_called}$=$true\,$\to$\,\textbf{VERIFY} (agent declared done but the verifier failed).
  \item Otherwise\,$\to$\,\textbf{REASON} (agent stopped voluntarily without claiming completion).
\end{enumerate}
This classification has no separate wall-time threshold; TIME is set by the \texttt{AgentTimeoutError} alone, which fires at the 900-second cap. A 20\% random sample was manually reviewed for consistency.

\section{Selected 50 Task IDs}
\label{app:tasks}
The 50 evaluation tasks (alphabetical):
\begin{flushleft}\scriptsize\raggedright
advanced-json-to-rfc4180-csv-converter,
advanced-poker-hand-classifier,
analyze-and-run-encoded-payload,
analyze-arm-shellcode-network-connections,
analyze-fen-with-stockfish,
apache-log-security-analyzer,
automate-blind-graph-mapping,
bash-ddos-traffic-analyzer,
bash-tree-diff-sync,
benchmark-gcc-opt-levels,
boot-debian-qemu-with-ssh-check,
build-arm64-qemu-linux-with-custom-message,
build-coq-from-source,
build-graphicsmagick-1-3-45,
build-grpc-user-profile-service,
build-nginx-1-24-production-server,
compare-lasso-ridge-elasticnet,
compare-lasso-ridge-gene-expression,
compute-best-chess-move-san,
configure-localhost-ssh-key-login,
consolidate-valid-prod-credentials,
count-claude-tokens-medical-papers,
count-unique-person-names-conll2003,
debug-bst-segfault-with-gdb,
decode-go-ctf-credentials,
decrypt-and-restore-backup-fragments,
detect-c-feature-flags,
diagnose-and-repair-broken-pip-installation,
find-invalid-blockchain-transactions,
fix-docker-python-dependency-conflicts,
fix-game-server-turn-race-condition,
fix-nameerrors-using-aliases-mapping,
fix-neural-net-weight-init,
fix-numpy-einsum-optimize-compatibility,
implement-lz77-file-compressor,
implement-tensor-parallel-matmul,
migrate-fortran-mcsim-to-gfortran,
migrate-make-to-cmake-build,
mongodb-sales-aggregation-engine,
repair-broken-shell-data-pipeline,
restore-broken-pip-installation,
sanitize-jinja2-ssti-templates,
simulate-2d-sampling-with-acceptance-stats,
solve-chess-mate-in-two,
solve-colossal-cave-350-score,
solve-escape-room-puzzle-server,
solve-train-shunting-puzzle,
train-fasttext-style-subword-embeddings,
train-fraud-detection-model,
train-sarsa-taxi-agent.
\end{flushleft}

\section{Failure Examples (One per Category)}
\label{app:examples}

\small
\raggedright
\begin{itemize}\itemsep2pt
  \item \textbf{REASON.}
  \texttt{train-fasttext-style-subword-embeddings} (goose/qwen):
  10 turns, 16.8K tokens, 237s; agent stops voluntarily.

  \item \textbf{VERIFY.}
  \texttt{simulate-2d-sampling-}\allowbreak\texttt{with-acceptance-stats} (openhands-sdk/qwen):
  18 turns, 272K tokens; \texttt{finish\_called=true} but tests fail.

  \item \textbf{TIME.}
  \texttt{detect-c-feature-flags} (goose/qwen):
  28 turns, 943s, \texttt{AgentTimeoutError}.

  \item \textbf{MAX\_TURNS.}
  \texttt{solve-colossal-cave-350-score} (goose/qwen):
  40 turns, 296s, no solution.

  \item \textbf{HANG.}
  \texttt{restore-broken-pip-installation} (goose/qwen):
  0 productive turns; harness exits via
  \texttt{NonZeroAgentExitCodeError} during sandbox setup.

  \item \textbf{ERROR.}
  \texttt{build-arm64-qemu-linux-}\allowbreak\texttt{with-custom-message} (openhands-sdk/qwen):
  6s, \texttt{DaytonaAuthorizationError} due to CPU quota;
  infrastructure failure.
\end{itemize}

\end{document}